\documentclass{article}
\usepackage{graphicx}
\usepackage{amssymb}
\usepackage{latexsym}
\usepackage{amsmath}
\usepackage{bbm}
\begin{document}

\title{Deep learning for source camera identification on mobile devices}
\author{David {Freire-Obreg\'on}\textsuperscript{1}, Fabio Narducci\textsuperscript{2}, Silvio Barra\textsuperscript{3}\\
and Modesto {Castrill\'on-Santana}\textsuperscript{1}\\
\textsuperscript{1}Universidad de Las Palmas de Gran Canaria, Spain \\
\textsuperscript{2}Universit\`{a} Parthenope di Napoli, Italy \\
\textsuperscript{3}Universit\`{a} degli Studi di Cagliari, Italy \\
} 

\maketitle

\begin{abstract}
In the present paper, we propose a source camera identification (SCI) method for mobile devices based on deep learning. Recently, convolutional neural networks (CNNs) have shown a remarkable performance on several tasks such as image recognition, video analysis or natural language processing. A CNN consists on a set of layers where each layer is composed by a set of high pass filters which are applied all over the input image. This convolution process provides the unique ability to extract features automatically from data and to learn from those features. Our proposal describes a CNN architecture which is able to infer the noise pattern of mobile camera sensors (also known as camera fingerprint) with the aim at detecting and identifying not only the mobile device used to capture an image (with a 98\% of accuracy), but also from which embedded camera the image was captured. More specifically, we provide an extensive analysis on the proposed architecture considering different configurations. The experiment has been carried out using the images captured from different mobile devices cameras (MICHE-I Dataset) and the obtained results have proved the robustness of the proposed method.
\end{abstract}

\section{Introduction}

The evolution of information and communication technologies (ICT) and the development of new mobile devices have provided to society with an unlimited number of online applications. 
Furthermore, the availability of these cost-effective, mobile and highly usable digital multimedia devices have made it possible to capture image and audio data without time, location and network related constraints.
This situation draws a new scenario in biometrics and forensics. One big concern is the growth of technology related felonies during the last decade, see~\cite{Lorang16}. Among these felonies, sexting, video voyeurism or revenge porn are the most common. Both,
sexting and revenge porn imply sending a sexually explicit material to a third party and the source device of this material is well known. However, video voyeurism requires hidden cameras (smartphones, nanny cameras, webcams or different kinds spy cameras) that secretly record or capture images of victims and disseminate those images remotely without the victims consent.

\begin{figure}[hbt]  
\begin{minipage}[t]{1\linewidth}
    \centering
    \includegraphics[scale=0.4]{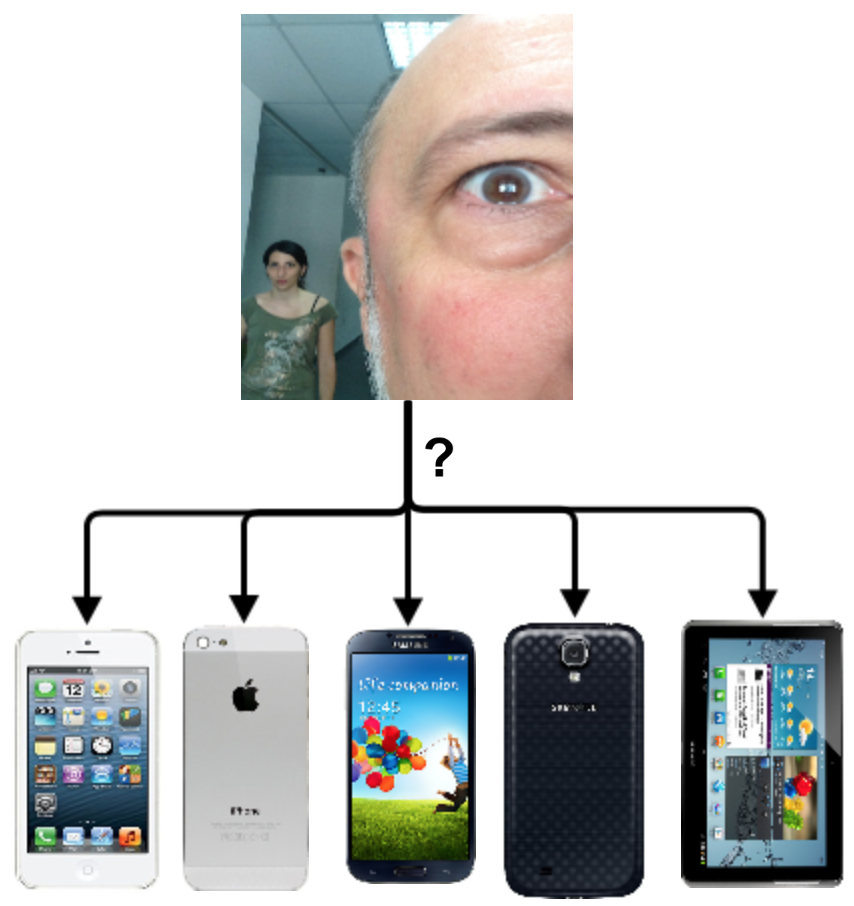}
      \end{minipage}
    \caption{Source camera identification on mobile devices tackles the problem to identify the device that captured an image. 
     \label{fig:objective}}
\end{figure}

The estimation of the victims ages or their identities is possible due to biometric techniques. In fact, the use of biometric recognition techniques has shown an excellent performance since the past century in several fields. Different biometric traits have been successfully applied in practical mobile applications including facial (\cite{Schroff2015}), fingerprint (\cite{Varol2015}), hand (\cite{Barra2017105}), iris (\cite{Abate201737, Abate201466}) or voice recognition (\cite{Baloul2012}). In this sense, law enforcement applications and the evolution of the technology have generated a huge interest in the usage of those biological traits for automated person recognition. Nonetheless, there are many situations where the biometric data is acquired under unconstrained environments and the quality of this data could be insufficient or particularly challenging for recognition (\cite{Neves2016}). When this situation occurs in crime scenes, forensic techniques can complement the lack of biometric information, see~\cite{Jain2015}. 

Source camera identification provides a mean to identify the exact device used for capturing an image (see Figure \ref{fig:objective}). In the video voyeurism felony, SCI does not extract any relevant information from the victim but it does from the perpetrator. Therefore, it answers how images were captured or which sensor captured the images. However, there is an important ground to be covered due to the fact that studies of SCI on new mobile devices are not sufficiently addressed in the literature.  

As a consequence, the major contributions of this study are as follows: 1) to tackle the SCI problem, 2) the design of an efficient CNN architecture, 3) the evaluation of different CNN configurations, and finally, 4) the successful application of the proposed approach to boost the SCI performance. 

The paper is organized in six sections. The next section looks at SCI related work. In section \ref{CNN_Arq}, the different CNN layers and the global architecture are described. The experimental setup as well as the classification experiments are reported in section \ref{sec:res}. Then, the CNN architecture discussion is addressed in section \ref{sec:discussion}. Finally, conclusions are drawn in section \ref{sec:concl}.

\begin{figure*}[t]  
\begin{minipage}[t]{1\linewidth}
    \centering
    \includegraphics[scale=0.45]{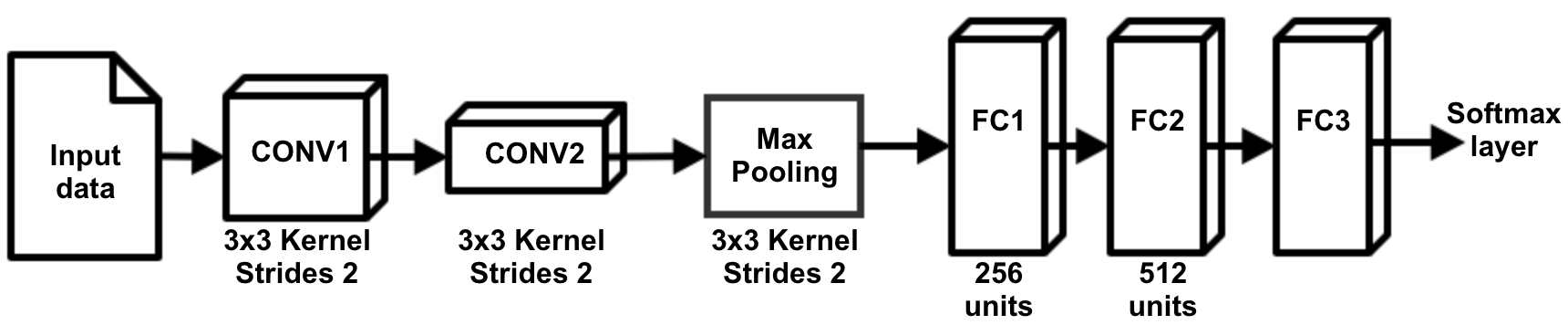}
      \end{minipage}
    \caption{The proposed CNN architecture for the SCI problem. It is composed by three main blocks: two convolutional layers ($CONV$), a max pooling layer (\textit{Max Pooling}) and three fully connected layers ($FC$).
     \label{fig:CNNArc}}
\end{figure*}

\section{State of the art}
\label{sec:soa}
In the previous section, SCI was defined as the process of determining which camera device has been used to capture an image. The information provided by the SCI technology is very helpful as evidence for legal issues, see~\cite{Casey11}. In this section, we address the way this problem has been tackled in the past.

The SCI approaches can be classified in two groups. The first group gathers all the methods that require to compute a model to identify a camera. Later, they evaluate a statistical proximity such as correlation between the image and the computed model. In this regard,~\cite{Choi06} demonstrated that it is possible to achieve a high rate of accuracy in the identification by measuring the intrinsic lens radial distortion of each camera. They rely on the fact that most of digital cameras are equipped with lenses having rather spherical surfaces, whose inherent radial distortions serve as unique fingerprints in the images. Then, they used parameters from aberration measurements to train and test a support vector machine (SVM) classifier. Another interesting work in the same line was proposed by~\cite{Chennamma10}. They defined the readout noise as an important intrinsic characteristic of a digital imaging sensor that cannot be removed. Thus, they proposed a study that measures readout noise of the sensor from an image using the mean-standard deviation plot in order to solve the SCI problem.

The second group is based on the use of methods that consider feature vector extraction and classical machine learning algorithms.~\cite{Filler08} introduced the sensor photo-response non-uniformity (PRNU) to solve the problem of digital cameras sensor identification. They defined the PRNU as the main component of a camera fingerprint, being a specific feature for each camera. Their classification process achieved an error rate of 11.2\%. Another interesting work was proposed by~\cite{Kharrazi04}. They developed a supervised learning approach based on features extracted from both, the spatial domain and the wavelet domain. They used a SVM to classify from five different cameras, obtaining error rates between 5\% and 22\%. The sensor pattern noise (SPN) based source camera identification technique has been also well established to solve the SCI problem. However,~\cite{Liu15} did an extensive evaluation among various ways of enhancing the SPN. They identified which enhancing methods offer some insights into the behavior of SPN for SCI and they achieved a roughly 15\% error rate. In this class of method, the work by ~\cite{lukas2006digital} also achieved interesting results at the cost of a high demanding computation though ~\cite{Cattaneo2017475} proposed a feasible distributed and scalable implementation of it.

Deep learning, in particular CNNs, have shown a quite good performance in several computer vision tasks such as  facial recognition (see \cite{Parkhi15}), pedestrian detection (see \cite{Angelova15}) or handwriting recognition (see \cite{Elleuch16}). Unlike the previous commented works based on feature extraction and classical machine learning algorithms, deep learning relies on the input data ability to drive their own feature extraction process. Concretely, a CNN is a complex computational model partially inspired by the human neural system that consists of a high number of interconnected nodes. These nodes are organized in multiple stacked layers performing a simple operation on the input. Then, after minimizing a cost function during the training process, the CNN is able to capture patterns in the input data.

Precisely, the present work focuses on designing a deep learning architecture in order to have an effective tool for SCI. The proposal is based on a CNN model, to provide good discrimination among different source cameras. The architecture as well as the hyperparameters of the presented approach builds a decision system to classify the given image into corresponding source category.

\section{CNN architecture}
\label{CNN_Arq}
As aforementioned, a CNN does not differ too much from the ordinary neural networks; they are made up by neurons that have parameters (weights and biases) which can be modified during the training process. Thus, each neuron receives a set of inputs and computes a set of operations all over the input data. Usually, this computation process is followed by a non-linearity operation. At the end of the process, the result feeds a loss function that has a key role in each neuron parameters update.
Contrary to the ordinary neural networks, the CNN only considers images as input. This kind of input allows to fit some specific properties of the architecture. The idea behind these properties is to reduce the amount of parameters in the network to make feasible the gradient computation and to preserve the assumption of locality. 

CNNs can be defined as a sequence of layers where each layer transforms a volume of inputs to a new volume of outputs through a differentiable functions. These layers are very specific to the task and they may differ to each other along the architecture. In this regard, we use three types of layers to build our CNN architecture: 1) convolutional, 2) pooling, and 3) fully-connected. Usually the selection of the architecture configuration is experimental, depends on how well the model behaves in order to achieve a given task. Then, the number of each layer type and the hyperparameter selection is tunned depending on the results of the model. This fine tune process of the proposed architecture is widely discussed in section \ref{sec:discussion} while in this section we detail the selected model shown in Figure \ref{fig:CNNArc}.

\subsection{Convolutional layer}
The parameters of each convolutional layer consist of a set of trainable kernels (also known as filters). As can be seen in Figure \ref{fig:CNNArc}, each of those kernels is small ($3\times3$ pixels) from a spatial perspective (height and width), but it extends through the input data volume. According to~\cite{Goodfellow16}, given a two dimensional image $I$ as input and a two dimensional kernel $K$, the formal computation of the convolutional layer can be described as in Equation \ref{eq:conv_eq}.

\begin{equation}
\label{eq:conv_eq}
S_{t}\left(i,j\right) = (K*I)(i,j) = \sum_{m=0}^{2}\sum_{n=0}^{2}I\left(i-m,j-n\right)K\left(m,n\right) 
\end{equation}

The generated feature map ($S_{t}(i,j)$) is the output of one kernel ($K$) applied to the previous layer ($S_{t-1}$) or the input image ($I$) in the case of the first layer. Then, each kernel is drawn across the entire previous layer, moving two pixels at a time due to the specifications provided for the proposed architecture in Figure \ref{fig:CNNArc}. Indeed, the \textit{strides} parameter is set to 2. As may be inferred, the two strides offset reduces the feature maps to a half of the related layer input data. 

Finally, each position result is an activation of the neuron and the output is collected in the feature map. The chosen activation function for this purpose is the Leaky Rectified Linear Unit (Leaky ReLU) function, see~\cite{Maas13}. The SCI problem is very sensitive to low level noise information. As can be seen in Equation \ref{eq:lrelu_eq}, unlike the regular Rectified Linear Unit (ReLU) function proposed by~\cite{HintonRelu10}, the Leaky ReLU allows a non-zero gradient when the unit is not active. This particularity provides more consistency of the benefit to address the task as shown in section \ref{sec:discussion}.

\begin{equation}
\label{eq:lrelu_eq}
f\left(x\right) = \mathbbm{1}\left(x<0\right)\left(\alpha x\right) + \mathbbm{1}\left(x>=0\right)\left(x\right)
\end{equation}

\subsection{Pooling layer}
Usually, pooling layers are inserted in-between successive convolutional layers. Their task is to reduce both, the number parameters and the computation complexity of the model. The way to achieve this is by progressively reducing the spatial size of the input representation. This is done in a similar way to the stride step defined in the convolutional layer. However, in this case, the maximum number in every subregion that the filter convolves around is extracted from the input volume. As can be seen in Figure \ref{fig:CNNArc}, the pooling layer is located between every convolutional and fully connected layers pair. In our architecture, the max pooling operation has a window size of $3\times3$. 

\subsection{Fully connected layer}
Finally three fully connected layers were considered at the end of the process. Basically, this is an ordinary neural network that takes an input volume (whatever the output of the pool layer preceding it is) and outputs an N dimensional vector where N is the number of classes that the architecture has been trained for. 

The fully-connected layers $FC1$ and $FC2$ have 256,
and 512 neurons respectively. In this case, the Leaky ReLU activation function is also
applied to the output of fully connected layer. The output of last fully connected layer $FC3$ feeds a softmax function.

To prevent overfitting, we have considered dropout, see~\cite{Hinton12}, as regularization technique. At each training stage, individual nodes between layers are ''dropped out'' of the CNN with probability $1-p$, or kept with a probability of $p$. Moreover, by droping out a random set of activations in a layer by setting them to zero, dropout decreases overfitting in the CNN. This technique also significantly improves the speed of training and seems to reduce neurons interactions, leading them to learn more robust features that better generalize to new data. In this proposed architecture, $FC1$ and $FC2$ are dropped out during the training process in order to avoid overfitting.

\begin{figure}[hbt]  
\begin{minipage}[t]{1\linewidth}
    \centering
    \includegraphics[scale=0.7]{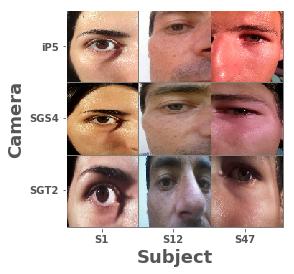}
      \end{minipage}
    \caption{MICHE-I Dataset. In the image we can appreciate three different subjects (S1, S12 and S47) from each device. Moreover, $iP5$ stands for iPhone5, $SGS4$ stands for Galaxy Samsung IV and $SGT2$ stands for Galaxy Tablet II respectively.
     \label{fig:MICHEDataset}}
\end{figure}

\section{Results}
\label{sec:res}
This section is divided into two subsections related to experimental issues: setup and results. The first subsection describes not only the considered dataset for our particular problem but also how we have prepared the data to feed the CNN. Then, the achieved results are summarized in the second subsection.

\subsection{Experimental setup}
We have used the MICHE-I Dataset for our experiments.~\cite{DeMarsico15} introduced this unique dataset of images taken from several mobile devices, especially for the purpose of development of mobile-devices-based forensic and biometric methods (\cite{InsightsMICHE},~\cite{Castrillón-Santana2017149}). The database is freely available for scientific purposes. Furthermore, the MICHE-I consists of over $3732$ images where three different devices where considered; iPhone5, Galaxy Samsung IV and Galaxy Tablet II. The authors gathered not only images of diverse indoor and outdoor scenes acquired under widely comparable conditions, but also images taken from different cameras (front and back camera) with the same device. Both devices, iPhone5 and Galaxy Samsung IV have two cameras, while the Galaxy Tablet II only has the back camera. In fact, this dataset provides us an interesting challenge by identifying both, the source of the image depending on the device model and the source of the image depending from which camera of the device it was taken. Figure~\ref{fig:MICHEDataset} shows images of three subjects for the three previously commented devices. 

However, for our purpose the dataset images were divided into $256$ smaller patches ($32\times32$) to fit the CNN model conditions. A bigger dataset that benefits the training process was assured by applying this non-overlapping patch division step. Furthermore, the training/testing division into two sets was done before the patch division step, so we ensure not to train and test with patches from the same image. 

For each experiment, train and test data are chosen randomly and the results are averaged after considering 10-fold cross validation. The experiments were conducted with a Nvidia Tesla K80 GPU. Several CNN configurations were tested looking at the minimum error rate, and the one described in the previous section achieves the best rates.

\subsection{Experimental results}

We conducted a set of experiments in order to validate the effectiveness of the selected CNN architecture. We built two different datasets for model-level and sensor-level camera identification. To build both datasets, we divide that MICHE-I depending on the model of the device and depending on the camera respectively:
\begin{enumerate}
\item The model-level camera experiment consisted on three output classes: iPhone5 (IP5), Galaxy Samsung IV (SG4) and Galaxy Tablet II (SGT2). 
\item The sensor-level camera experiment consisted on five output classes: iPhone5 front camera (IP5\_F), iPhone5 back camera (IP5\_B), Galaxy Samsung IV front camera (SG4\_F), Galaxy Samsung IV back camera (SG4\_B) and Galaxy Tablet II front camera (SGT2\_F).
\end{enumerate}

\begin{figure}[hbt]  
\begin{minipage}[t]{1\linewidth}
    \centering
    \includegraphics[scale=0.30]{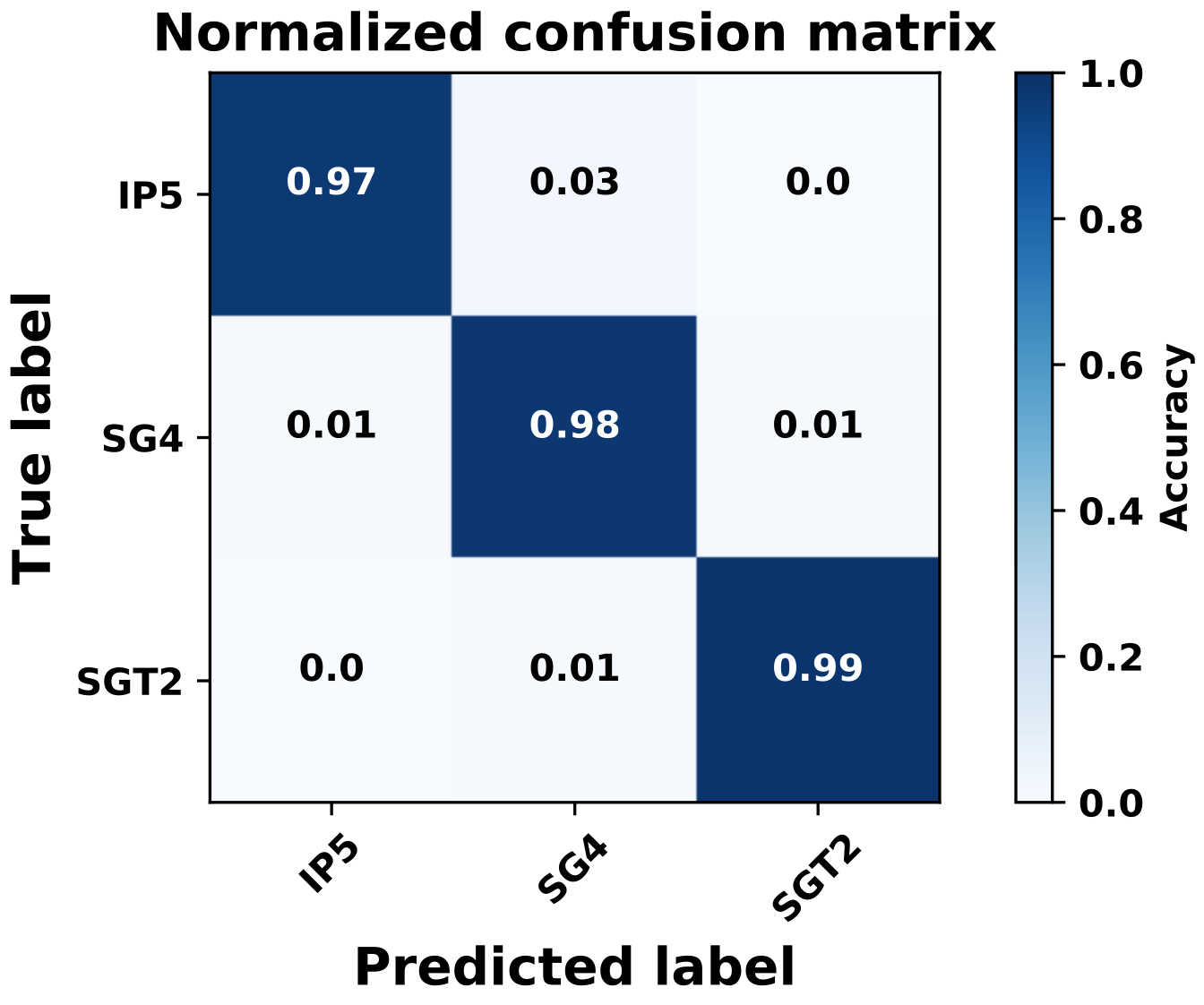}
      \end{minipage}
    \caption{Confusion matrix of the model-level experiment for the SCI task. The different device models are denoted as iPhone5 (IP5), Galaxy Samsung IV (SG4) and Galaxy Tablet II (SGT2) respectively.
     \label{fig:brandexp}}
\end{figure}

On the one hand, Figure~\ref{fig:brandexp} shows the confusion matrix obtained on the test set when the model-level camera experiment was considered. The precision for this experiment is as high as $98.1\%$. As shown by the confusion matrix, the proposed CNN exhibits a good performance to identify the device from which the image was obtained.

\begin{figure}[hbt]  
\begin{minipage}[t]{1\linewidth}
    \centering
    \includegraphics[scale=0.22]{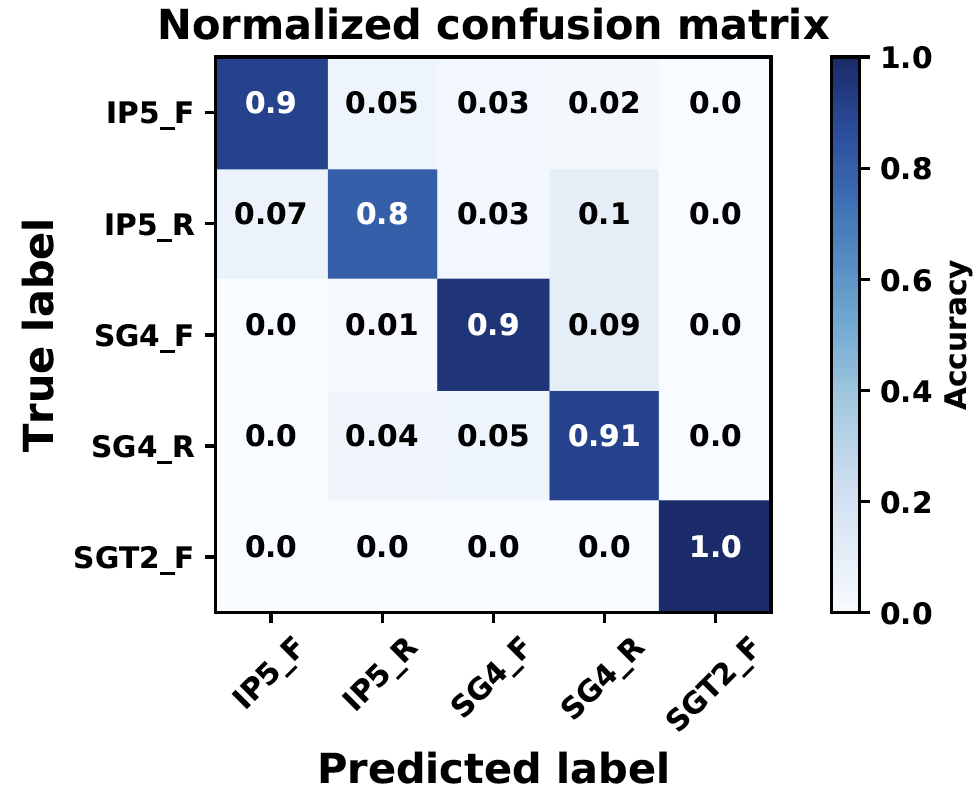}
      \end{minipage}
    \caption{Confusion matrix of the sensor-level experiment for the SCI task. The different camera models are denoted as iPhone5 front camera (IP5\_F), iPhone5 back camera (IP5\_B), Galaxy Samsung IV front camera (SG4\_F), Galaxy Samsung IV back camera (SG4\_B) and Galaxy Tablet II front camera (SGT2\_F) respectively.
     \label{fig:camexp}}
\end{figure}

On the other hand, Figure~\ref{fig:camexp} shows the confusion matrix obtained on the test set when we conducted the sensor-level camera experiment. In this experiment, the overall accuracy on the test set is $91.1\%$. As shown by the confusion matrix, the proposed CNN is also quite good at discriminating between the different camera models. However it is possible to note that it is more difficult to distinguish not only between cameras models of the same manufacturer but also cameras located at the same position of the device. For instance, it is more complicated to distinguish images between $IP5\_F$ and $IP5\_R$ (same device) than distinguishing images taken from any of the $IP5$ cameras and the $SGT2$ camera. 

Comparing the results obtained from both experiments, it can be inferred that the performance is quite good in the case of identifying the source device of the image, whereas it slightly degrades when there are multiple cameras belonging to the same device manufacturer. This is due, as it can be observed in~\cite{DeMarsico15} work, to the strong feature similarity of some camera models from the same manufacturer.

\begin{figure}[hbt]  
\begin{minipage}[t]{1\linewidth}
    \centering
    \includegraphics[scale=0.34]{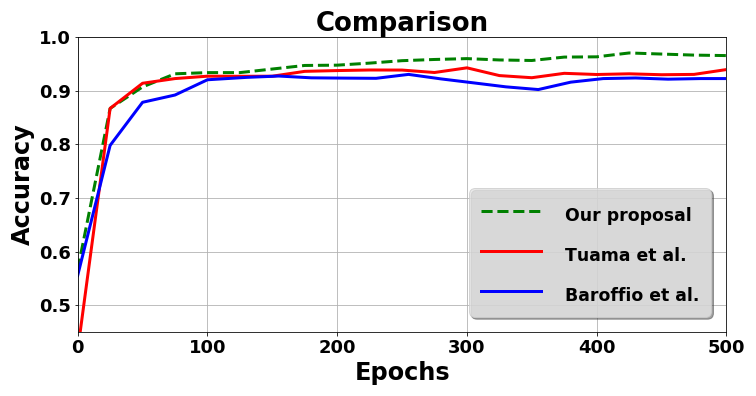}
      \end{minipage}
    \caption{Comparison between our proposal and two different approaches considering the MICHE-I dataset on the model-based detection.
     \label{fig:comp}}
\end{figure}

In order to compare, we have selected two remarkable works that face the SCI problem considering deep learning. None of them have been tested on a mobile devices dataset before and both have been used to address the model-based detection problem. On the one hand,~\cite{Baroffio16} proposed a three convolutional and two fully connected layers architecture. As can be seen in Figure \ref{fig:comp}, their complex architecture achieves for an identical experimental set, a $92.3\%$ of accuracy on the MICHE-I dataset. In spite of the high number of feature maps generated during the pipeline, we believe that considering three convolutional layers (plus a pooling layer after the first two convolutional layers) is downsizing considerably the sensitive information. On the other hand,~\cite{Tuama16} proposed an architecture also based on three convolutional layers, but unlike~\cite{Baroffio16}, they applied a single pooling layer after the convolutional layers. This second architecture achieves a $93.4\%$ of accuracy on the model-based detection. Figure \ref{fig:comp} evidences that our approach outperforms both proposals on the MICHE-I dataset.

\section{Discussion}
\label{sec:discussion}

In section \ref{CNN_Arq}, we described the CNN architecture. The selection of the architecture topology as well as the hyperparameters tunning are not a trivial tasks, they require a deeper analysis and to understand both, the theoretical and the practical issues. This happens because the CNN configuration depends on the data type we are dealing with. For instance, the data can vary by size, complexity of the images or the task we are solving. In this section, we argue about the different elements that support the selection of the proposed CNN architecture.

\begin{figure}[hbt]  
\begin{minipage}[t]{1\linewidth}
    \centering
    \includegraphics[scale=0.34]{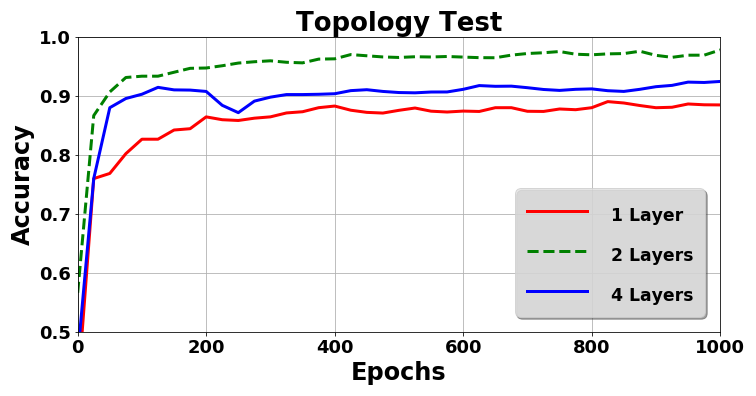}
      \end{minipage}
    \caption{The topology experiment. Three different topologies are shown in this graph. The 1 convolutional-layer-approach tends to underfit, whereas the 4 convolutional-layers-approach tends to overfit. The optimal accuracy occurs when 2 convolutional layers are considered.
     \label{fig:layers_disc}}
\end{figure}

In the first experiment, we studied the effect of applying different convolution layers to the architecture. To this end, we modified the number of convolutional layers at each level of the standard CNN defined in section \ref{CNN_Arq}. Usually, in the first few layers of a CNN the network we can identify lines and corners. Then, these patterns are passed down through the neural net, starting to recognize more complex features as we get deeper as~\cite{Zeiler13} stated. However, the SCI problem is not about recognizing complex structures but to identify the digital signals that differs from each camera. As it is shown in Figure~\ref{fig:layers_disc}, this kind of signal is very sensitive and rapidly tends to overfit (in the case of more than two convolutional layers) or to underfit in the case of just one convolutional layer. In this case, the two convolutional layers topology has shown the best performance.

\begin{figure}[hbt]  
\begin{minipage}[t]{1\linewidth}
    \centering
    \includegraphics[scale=0.34]{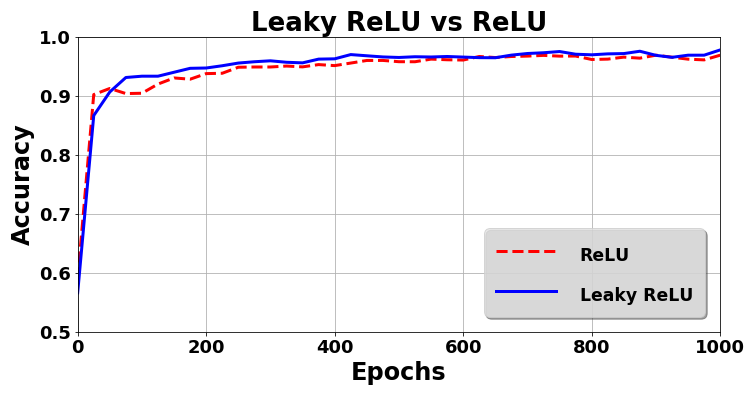}
      \end{minipage}
    \caption{The activation function experiment. Leaky ReLU assures to keep alive the network nodes. As a consequence, more nodes stay active working on the task and provides a slightly higher accuracy.
     \label{fig:activ_disc}}
\end{figure}

Secondly, the activation function effect is addressed. These functions introduce non linearities to the output of the CNN models. In spite of the existence of several activation functions (e.g. sigmoid, tanh, ReLU, etc.), only two of them were considered in our experiments. Krizhevsky et al. demonstrated that ReLU not only outperforms \emph{tanh} by six times in convergence terms, but also involves simpler operations compared to \emph{tanh} or \emph{sigmoid} as~\cite{Krizhevsky12} stated. Unfortunately, ReLU units can be fragile during the training step. Therefore, a large gradient flowing through a ReLU node could cause the weights to update in such a way that the node will never activate again. This problem can lead to a scenario where a 25-35\% can be "dead". Leaky ReLU tries to solve this issue by adding a small slope to the negative values keeping all the positive properties that ReLU introduced. As can be appreciated in Figure~\ref{fig:activ_disc}, there is not a significant difference in terms of accuracy. However, Leaky ReLU exhibits a better rate (i.e. $2\%$-$4\%$ of improvement). 

\begin{figure}[hbt]  
\begin{minipage}[t]{1\linewidth}
    \centering
    \includegraphics[scale=0.34]{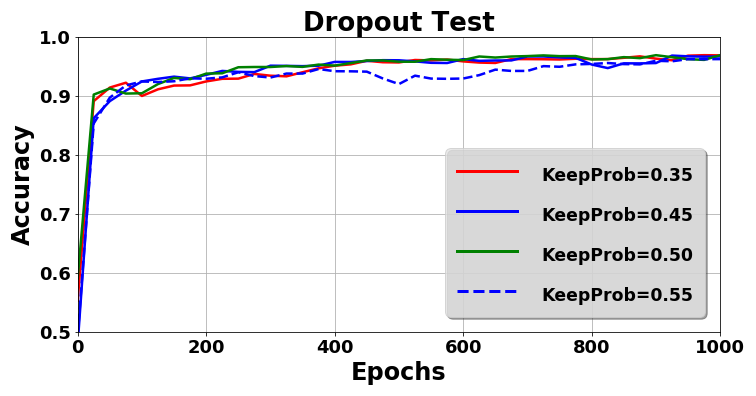}
      \end{minipage}
    \caption{The dropout hyperparameter study. Different probabilities to keep nodes are tested; $0.35\%$, $0.45\%$, $0.5\%$ and $0.55\%$ respectively.  
     \label{fig:dropout_disc}}
\end{figure}

The last experiment studies the effect of varying parameters of the regularization algorithm. A new hyperparameter was introduced when we described the dropout technique to prevent overfitting, the probability $p$ of retaining each layer node. While training, dropout is implemented by only keeping a node active with this probability $p$, or setting it to zero otherwise. We conducted an experiment applying different values for the hyperparameter $p$. ~\cite{Srivastava14} claimed that depending on the input, typical values of $p$ are in the range $0.5$ to $0.8$. They also argued that a large $p$ may not produce enough dropout to prevent overfitting. As can be seen in Figure~\ref{fig:dropout_disc}, the best accuracy is achieved when the value of $p$ is $0.5$.  

These three important issues are some of many others that have been addressed during the development of the proposed CNN architecture. As aforementioned, several architectures were tested. Consequently, our decision guide process was based on experimental results. However, literature also supports our decisions. For instance, the convolutional layers uses $3\times3$ kernels to process the input data.~\cite{Simonyan2014} have shown that $3\times3$ convolution kernels prove to be efficient because they not only make the decision function more discriminative but they also reduce exponentially the number of parameters when depth is increased.

\section{Conclusion}
\label{sec:concl}
In this paper, we presented a study to address the SCI problem on mobile devices based on deep learning. For this reason, we have conducted several experiments considering CNNs on a dataset of images acquired from mobile devices. The contribution represents an interesting challenge since there is not much research on this kind of dataset for SCI. 

As a consequence, our contribution operates from small patches and identifies not only the device from which the image was taken, but also the camera within the device that took it. Indeed, the model is sufficiently complex to effectively distinguish between both problems, a $98.1\%$ of accuracy for identifying the device manufacturer and $91.1\%$ of accuracy for identifying the exact camera. 

We have also discussed how varying the network topology or how tuning the hyperparameters can affect the model accuracy. In this sense, scalability has been evaluated and increasing the number of convolutional layers seems not to improve a simpler model due to the size of the input data.

\bibliographystyle{apalike}
\bibliography{refs}

\begin{thebibliography}{}

\bibitem[Abate et~al., 2017]{Abate201737}
Abate, A., Barra, S., Gallo, L., and Narducci, F. (2017).
\newblock Kurtosis and skewness at pixel level as input for som networks to
  iris recognition on mobile devices.
\newblock {\em Pattern Recognition Letters}, 91:37--43.
\newblock cited By 0.

\bibitem[Abate et~al., 2014]{Abate201466}
Abate, A., Nappi, M., Narducci, F., and Ricciardi, S. (2014).
\newblock Fast iris recognition on smartphone by means of spatial histograms.
\newblock {\em Lecture Notes in Computer Science (including subseries Lecture
  Notes in Artificial Intelligence and Lecture Notes in Bioinformatics)},
  8897:66--74.
\newblock cited By 6.

\bibitem[Angelova et~al., 2015]{Angelova15}
Angelova, A., Krizhevsky, A., Vanhoucke, V., Ogale, A., and Ferguson, D.
  (2015).
\newblock Real-time pedestrian detection with deep network cascades.
\newblock In {\em Proceedings of the British Machine Vision Conference
  ({BMVC})}, pages 1--12.

\bibitem[Baloul et~al., 2012]{Baloul2012}
Baloul, M., Cherrier, E., and Rosenberger, C. (2012).
\newblock Challenge-based speaker recognition for mobile authentication.
\newblock In {\em 2012 BIOSIG - Proceedings of the International Conference of
  Biometrics Special Interest Group (BIOSIG)}, pages 1--7.

\bibitem[Baroffio et~al., 2016]{Baroffio16}
Baroffio, L., Bondi, L., Bestagini, P., and Tubaro, S. (2016).
\newblock Camera identification with deep convolutional networks.
\newblock {\em CoRR}, abs/1603.01068.

\bibitem[Barra et~al., 2017]{Barra2017105}
Barra, S., {De Marsico}, M., Nappi, M., Narducci, F., and Riccio, D. (2017).
\newblock Mohab: Mobile hand-based biometric recognition.
\newblock {\em Lecture Notes in Computer Science (including subseries Lecture
  Notes in Artificial Intelligence and Lecture Notes in Bioinformatics)}, 10232
  LNCS:105--115.
\newblock cited By 0.

\bibitem[Casey, 2011]{Casey11}
Casey, E. (2011).
\newblock {\em Digital Evidence and Computer Crime: Forensic Science,
  Computers, and the Internet}.
\newblock Academic Press, 3rd edition.

\bibitem[Castrill\'on-Santana et~al., 2017]{Castrillón-Santana2017149}
Castrill\'on-Santana, M., {De Marsico}, M., Nappi, M., Narducci, F., and
  Proença, H. (2017).
\newblock Mobile iris challenge evaluation ii: Results from the icpr
  competition.
\newblock pages 149--154.
\newblock cited By 0.

\bibitem[Cattaneo et~al., 2017]{Cattaneo2017475}
Cattaneo, G., Petrillo, U., Nappi, M., Narducci, F., and Roscigno, G. (2017).
\newblock An efficient implementation of the algorithm by lukáš et al. on
  hadoop.
\newblock {\em Lecture Notes in Computer Science (including subseries Lecture
  Notes in Artificial Intelligence and Lecture Notes in Bioinformatics)}, 10232
  LNCS:475--489.
\newblock cited By 0.

\bibitem[Chennamma and Rangarajan, 2010]{Chennamma10}
Chennamma, H.~R. and Rangarajan, L. (2010).
\newblock Source camera identification based on sensor readout noise.
\newblock {\em IJDCF}, 2(3):28--42.

\bibitem[Choi et~al., 2006]{Choi06}
Choi, K.~S., Lam, E., and Wong, K. (2006).
\newblock Automatic source camera identification using the intrinsic lens
  radial distortion.
\newblock In {\em Proc. SPIE, Digital Photography II}, volume~24, pages
  11551--11565. Optics Express.

\bibitem[{De Marsico} et~al., 2018]{InsightsMICHE}
{De Marsico}, M., Nappi, M., Narducci, F., and Proença, H. (2018).
\newblock Insights into the results of miche i - mobile iris challenge
  evaluation.
\newblock {\em Pattern Recognition}, 74(Supplement C):286 -- 304.

\bibitem[{De Marsico} et~al., 2015]{DeMarsico15}
{De Marsico}, M., Nappi, M., Riccio, D., and Wechsler, H. (2015).
\newblock Mobile iris challenge evaluation (miche)-i, biometric iris dataset
  and protocols.
\newblock {\em Pattern Recogn. Lett.}, 57(C):17--23.

\bibitem[Elleuch and Kherallah, 2016]{Elleuch16}
Elleuch, M. and Kherallah, M. (2016).
\newblock An improved arabic handwritten recognition system using deep support
  vector machines.
\newblock {\em Int. J. Multimed. Data Eng. Manag.}, 7(2):1--20.

\bibitem[Filler et~al., 2008]{Filler08}
Filler, T., Fridrich, J.~J., and Goljan, M. (2008).
\newblock Using sensor pattern noise for camera model identification.
\newblock In {\em ICIP}, pages 1296--1299. IEEE.

\bibitem[Goodfellow et~al., 2016]{Goodfellow16}
Goodfellow, I., Bengio, Y., and Courville, A. (2016).
\newblock {\em Deep Learning}.
\newblock MIT Press.

\bibitem[Hinton et~al., 2012]{Hinton12}
Hinton, G.~E., Srivastava, N., Krizhevsky, A., Sutskever, I., and
  Salakhutdinov, R. (2012).
\newblock Improving neural networks by preventing co-adaptation of feature
  detectors.
\newblock {\em CoRR}, abs/1207.0580.

\bibitem[Jain and Ross, 2015]{Jain2015}
Jain, A.~K. and Ross, A. (2015).
\newblock Bridging the gap: from biometrics to forensics.
\newblock {\em Philosophical Transactions of the Royal Society of London B:
  Biological Sciences}, 370(1674).

\bibitem[Kharrazi et~al., 2004]{Kharrazi04}
Kharrazi, M., Sencar, H.~T., and Memon, N.~D. (2004).
\newblock Blind source camera identification.
\newblock In {\em ICIP}, pages 709--712. IEEE.

\bibitem[Krizhevsky et~al., 2012]{Krizhevsky12}
Krizhevsky, A., Sutskever, I., and Hinton, G.~E. (2012).
\newblock Imagenet classification with deep convolutional neural networks.
\newblock In {\em Advances in Neural Information Processing Systems}, page~9.

\bibitem[Liu et~al., 2015]{Liu15}
Liu, B., Wei, X., and Yan, J. (2015).
\newblock Enhancing sensor pattern noise for source camera identification: An
  empirical evaluation.
\newblock In {\em Proceedings of the 3rd ACM Workshop on Information Hiding and
  Multimedia Security}, IH\&\#38;MMSec '15, pages 85--90, New York, NY, USA.
  ACM.

\bibitem[Lorang et~al., 2016]{Lorang16}
Lorang, M.~R., McNiel, D.~E., and Binder, R.~L. (2016).
\newblock Minors and sexting: Legal implications.
\newblock {\em Journal of the American Academy of Psychiatry and the Law
  Online}, 44(1):73--81.

\bibitem[Lukas et~al., 2006]{lukas2006digital}
Lukas, J., Fridrich, J., and Goljan, M. (2006).
\newblock Digital camera identification from sensor pattern noise.
\newblock {\em IEEE Transactions on Information Forensics and Security},
  1(2):205--214.

\bibitem[Maas et~al., 2013]{Maas13}
Maas, A.~L., Hannun, A.~Y., and Ng, A.~Y. (2013).
\newblock Rectifier nonlinearities improve neural network acoustic models.
\newblock In {\em in ICML Workshop on Deep Learning for Audio, Speech and
  Language Processing}, pages 13--19.

\bibitem[Nair and Hinton, 2010]{HintonRelu10}
Nair, V. and Hinton, G.~E. (2010).
\newblock Rectified linear units improve restricted boltzmann machines.

\bibitem[Neves et~al., 2016]{Neves2016}
Neves, J., Narducci, F., Barra, S., and Proen{\c{c}}a, H. (2016).
\newblock Biometric recognition in surveillance scenarios: a survey.
\newblock {\em Artificial Intelligence Review}, 46(4):515--541.

\bibitem[Parkhi et~al., 2015]{Parkhi15}
Parkhi, O.~M., Vedaldi, A., and Zisserman, A. (2015).
\newblock Deep face recognition.
\newblock In {\em Proceedings of the British Machine Vision Conference
  ({BMVC})}, pages 1--12.

\bibitem[Schroff et~al., 2015]{Schroff2015}
Schroff, F., Kalenichenko, D., and Philbin, J. (2015).
\newblock Facenet: A unified embedding for face recognition and clustering.
\newblock In {\em 2015 IEEE Conference on Computer Vision and Pattern
  Recognition (CVPR)}, pages 815--823.

\bibitem[Simonyan and Zisserman, 2014]{Simonyan2014}
Simonyan, K. and Zisserman, A. (2014).
\newblock Very deep convolutional networks for large-scale image recognition.
\newblock In {\em Proceedings of the International Conference on Learning
  Representations ({ICLR})}, pages 1--14.
\newblock cite arxiv:1409.1556.

\bibitem[Srivastava et~al., 2014]{Srivastava14}
Srivastava, N., Hinton, G., Krizhevsky, A., Sutskever, I., and Salakhutdinov,
  R. (2014).
\newblock Dropout: A simple way to prevent neural networks from overfitting.
\newblock {\em J. Mach. Learn. Res.}, 15(1):1929--1958.

\bibitem[Tuama et~al., 2016]{Tuama16}
Tuama, A., Comby, F., and Chaumont, M. (2016).
\newblock Camera model identification with the use of deep convolutional neural
  networks.
\newblock In {\em 2016 IEEE International Workshop on Information Forensics and
  Security (WIFS)}, pages 1--6.

\bibitem[Yıldırım and Varol, 2015]{Varol2015}
Yıldırım, N. and Varol, A. (2015).
\newblock Android based mobile application development for web login
  authentication using fingerprint recognition feature.
\newblock In {\em 2015 23nd Signal Processing and Communications Applications
  Conference (SIU)}, pages 2662--2665.

\bibitem[Zeiler and Fergus, 2013]{Zeiler13}
Zeiler, M.~D. and Fergus, R. (2013).
\newblock Visualizing and understanding convolutional networks.
\newblock {\em CoRR}, abs/1311.2901.

\end{thebibliography}

\end{document}